# Compositional Planning Using Optimal Option Models

**David Silver**  D.SILVER@CS.UCL.AC.UK
**Kamil Ciosek**  K.CIOSEK@CS.UCL.AC.UK
Department of Computer Science, CSML, University College London, Gower Street, London WC1E 6BT.

## Abstract

In this paper we introduce a framework for *option model composition*. Option models are temporal abstractions that, like macro-operators in classical planning, jump directly from a start state to an end state. Prior work has focused on constructing option models from primitive actions, by *intra-option* model learning; or on using option models to construct a value function, by *inter-option* planning. We present a unified view of intra- and inter-option model learning, based on a major generalisation of the Bellman equation. Our fundamental operation is the recursive composition of option models into other option models. This key idea enables *compositional* planning over many levels of abstraction. We illustrate our framework using a dynamic programming algorithm that simultaneously constructs optimal option models for multiple subgoals, and also searches over those option models to provide rapid progress towards other subgoals.

## 1 Introduction

Classical planning algorithms make extensive use of temporal abstraction to construct high-level chunks of useful knowledge (Amarel, 1968; Sacerdoti, 1975; Korf, 1985; Laird et al., 1986). They are typically provided with a set of primitive planning operators as inputs. These are then composed together into *macro-operators*: open-loop sequences of planning operators. Macro-operators jump directly from an initial state to the outcome state that would result from following the sequence, without having to execute the intermediate operators. Macro-operators can themselves be composed together into more abstract operators, allowing planning to take place at a much more abstract level.

Macro-operators can be thought of as building blocks of knowledge, which can be combined together into more abstract knowledge. Powered by this knowledge, the path to the goal can often be found in a small number of high-level planning operations, even when the path is composed of thousands of primitive actions.

In Markov Decision Processes (MDPs), the outcome of an action may be stochastic. An open-loop sequence does not capture the contingencies that can arise as a result of each intermediate action. Instead, a closed-loop policy, which maps states to actions, can respond to each particular situation as it arises. A closed-loop policy that is followed for some number of steps, and stops according to a termination condition that also depends on the state, is known as an *option* (Sutton et al., 1999). An *option model* describes the distribution of outcome states that would result from following the option (Sutton, 1995). Option models are the stochastic analogue of macro-operators: they jump directly from initial state to outcome, without having to execute the intermediate actions. Option models can also be composed together into more abstract option models (Precup et al., 1998). Option models thus provide basic building blocks for compositional knowledge in general MDPs.

However, prior work on planning with options has been restricted to shallow hierarchies. Option models are either *constructed* from primitive actions, in an approach known as *intra-option* model learning; or they are *used* to compute a value function, in an approach known as *inter-option* (or SMDP) planning (Sutton et al., 1999). Although these steps are sometimes combined, they are typically combined in two stages: first constructing the option models without using them; and then using the option models without changing them. In both stages, the planning operators are fixed.

In this paper we focus explicitly on *compositional planning*: the multi-level composition of option models. Each option model is both constructed (intra-option) and used (inter-option). It is constructed from lower-level option models, so as to maximise progress towards a given subgoal. It may also be used to com-pose higher-level option models. As soon as an option model has been created, it can be used to construct other option models. As a result, the set of planning operators improves dynamically, providing longer and more purposeful jumps as planning proceeds.

Our approach is based on a major generalisation of the Bellman equation along four dimensions. First, we provide a recursive relationship between state probabilities as well as between rewards. Second, we compose over options rather than primitive actions. Third, we generalise from the overall goal of maximising total reward, to any given subgoal. Fourth, we optimise over termination conditions as well as policies.

Several of these dimensions have been partially explored by prior work. First, Sutton et al. (1995, 1999, Section 5) developed a Bellman expectation equation for state probabilities, but this work was restricted to Markov reward processes without actions (Sutton, 1995) or to fixed policies without control (Sutton et al., 1999). We present a Bellman *optimality* equation for state probabilities in Markov decision processes, including actions and control. Second, Precup *et al.* (1998) provided Bellman equations for composing option models into policies, but not into options. Our framework constructs both policies and termination conditions, so that we can compose option models into other option models – a crucial step for compositional planning. Third, Sutton *et al.* (1999, Section 7) defined optimal options with respect to a given subgoal and termination condition, and *suggested* the existence of a corresponding Bellman optimality equation. We *define* this Bellman optimality equation, and also extend to the case when neither, either, or both the policy and termination condition are specified. *No prior work has considered the state probabilities associated with Bellman optimality equations.* Without knowledge of these state probabilities, it is not possible to jump directly to the outcome of an optimal option. Our approach to compositional planning is built directly on this knowledge, so as to build abstract macro-operators that can jump from one state directly to a distant state.

The Bellman optimality equation gives rise to important planning methods such as value iteration. Similarly, we use our generalised Bellman equation to derive a compositional planning algorithm, which simultaneously and recursively constructs the optimal option model for multiple subgoals, including the overall goal as a special case. We prove that this algorithm converges to optimal option models for all subgoals, including the optimal policy.

The options framework is agnostic about the source of the options, and does not commit to any particular algorithm for their construction. However, several other approaches to hierarchical reinforcement learning have been proposed, based on samples from an unknown MDP. These architectures, including Dietterich's MAXQ (2000), and Parr and Russell's HAMs (1997; 2002), do construct the solution to one subproblem from the solution to other subproblems. However, these architectures are not directly applicable to compositional planning, where the MDP is known rather than sampled. By focusing on planning with known models, we develop a sound theoretical framework for compositional planning, based on the generalised Bellman equation. This work can be viewed as a bridge between the generality of options, and the compositional construction algorithms used by architectures such as MAXQ.

We illustrate our approach on two well-known benchmark problems: hierarchical path planning and the Tower of Hanoi. Both problems have been extensively studied using classical planning approaches. In both problems, planning directly with primitive operators (e.g. using value iteration) requires computation time that is exponential in the problem size, whereas algorithms based on compositions of macro-operators (e.g. Jonsson 2009) can solve these problems in polynomial time. Unfortunately, classical planning approaches do not generalise to stochastic planning problems. In contrast, our compositional planning algorithm can solve both deterministic *and* stochastic variants of these problems in a polynomial number of iterations.

## 2 Background

An MDP is defined by a set of $n$ states $\mathcal{S}$, a set of actions $\mathcal{A}$, action transition matrices $P^a$ and action reward vectors $R^a$ for each action $a \in \mathcal{A}$, and a discount factor $0 \leq \gamma < 1$. Each component of the action transition matrix $P^a_{ss'}$ is the discounted probability of next state $s'$ given that action $a$ was selected in state $s$, $P^a_{ss'} = \gamma \, Pr(s_{t+1} = s' \mid s_t = s, a_t = a)$. Each component of the action reward vector $R^a_s$ is the expected reward given that action $a$ was selected in state $s$, $R^a_s = \mathbb{E}[r_{t+1} \mid s_t = s, a_t = a]$. The discount factor can be viewed as a chance of *exiting* to an absorbing terminal state with probability $1 - \gamma$. The discounted probability $P^a_{ss'}$ can be interpreted as the probability of reaching state $s'$ without exiting.

A *policy* $\pi(s,a)$ is the probability of selecting action $a$ given state $s$, $\pi(s,a) = Pr(a_t = a \mid s_t = s)$. The *value function*, $V^\pi(s)$, is the expected total reward from state $s$ when following policy $\pi$, $V^\pi(s) = \mathbb{E}[r_{t+1} + \gamma r_{t+2} + ... \mid s_t = s, \pi]$. The *optimal value function* $V^*(s)$ and *optimal action value function* $Q^*(s,a)$ are the maximum achievable value and action value that can be achieved by any policy, $V^*(s) = \max_\pi V^\pi(s)$. An *optimal policy* $\pi^*(s,a)$ is any policy that achieves the optimal value function.

The optimal value function obeys a recursive relationship: the *Bellman optimality equation*, $V^*(s) = \max_a R_s^a + \sum_{s'} P_{ss'}^a V^*(s')$. The optimal value function is the unique fixed point of this equation, and can be found by turning the Bellman optimality equation into an iterative update, $V_{k+1}(s) \leftarrow \max_a R_s^a + \sum_{s'} P_{ss'}^a V_k(s')$. This algorithm is called *value iteration* (Bellman, 1957).

An *option* $o = \langle \pi, \beta \rangle$ is an extended behaviour or macro-action that combines a policy $\pi(s,a)$ with a *termination condition* $\beta(s)$ giving the probability that the option will stop in state $s$. We assume that options can be initiated from all states. Primitive actions are options: they can be represented by a policy that deterministically selects that action, and a termination condition that stops with probability 1. We denote the set of all policies by $\Pi$ and the set of all termination conditions by $\mathcal{B}$. An *option model* comprises an *option transition matrix* $P^o$ and an *option reward vector* $R^o$. Each component $R_s^o$ is the expected total reward given that option $o$ was executed from state $s$, $R_s^o = \mathbb{E}[r_{t+1} + \gamma r_{t+2} + ... + \gamma^{\tau-1} r_{t+\tau} \mid s_t = s, o]$, where $\tau$ is the random variable for the duration of option $o$. Each component $P_{ss'}^o$ is the probability of terminating in state $s'$ given that option $o$ was executed from state $s$, discounted by the total duration of the option, $P_{ss'}^o = \sum_{\tau=1}^\infty \gamma^\tau Pr(\tau, s_{t+\tau} = s' \mid s_t = s, o)$. This can be interpreted as the probability of option $o$ terminating in $s'$ without exiting.

## 3 Models

Informally, a model is a stochastic mapping from state to state, combined with the reward accumulated along the way. Applying a model to a state results in a distribution over outcome states, and an expected reward. To compose models together, we apply a second model to this outcome distribution and expected reward, and arrive at a new state distribution and reward. We now formalise these ideas, following Sutton (1995).

We define a *rasp* (reward and state probabilities), $[r \mid p]$, to be a $1 \times (1+n)$ row vector, where $n = |\mathcal{S}|$, $r$ is a scalar reward, and $p$ is a $1 \times n$ row vector representing a discounted probability distribution $p$ over states in $\mathcal{S}$. We use $\mathbf{s}_s$ to denote the deterministic rasp that is in state $s$ with probability 1, and has a reward component of zero; we shorten to $\mathbf{s}$ when there is no ambiguity. Rasps are ordered by their reward components, $[r_1 \mid p_1] \leq [r_2 \mid p_2]$ if and only if $r_1 \leq r_2$. A *model* is a transformation from rasp to rasp. Formally, a model $\left[\begin{array}{c|c} 1 & 0 \\ \hline R & P \end{array}\right]$ is a $1+n \times 1+n$ block matrix containing an $n \times 1$ reward vector $R$ and an $n \times n$ transition matrix $P$. This block matrix notation for models and block vector notation for rasps are known as *homogeneous coordinates* (Sutton, 1995). To *compose* two models together, we multiply their homogeneous coordinates,

$$\left[\begin{array}{c|c} 1 & 0 \\ \hline R_1 & P_1 \end{array}\right] \left[\begin{array}{c|c} 1 & 0 \\ \hline R_2 & P_2 \end{array}\right] = \left[\begin{array}{c|c} 1 & 0 \\ \hline R_1 + P_1 R_2 & P_1 P_2 \end{array}\right] \quad (1)$$

Similarly, to compose a rasp and a model, we again multiply their homogeneous coordinates,

$$[r \mid p] \left[\begin{array}{c|c} 1 & 0 \\ \hline R & P \end{array}\right] = [r + pR \mid pP]. \quad (2)$$

To aid readability, homogenous matrices and vectors are denoted by boldface letters, e.g $\mathbf{x}$, $\mathbf{M}$ and $\mathcal{M}$ for rasps, models and model sets respectively. Sequences of compositions are best understood by reading left to right, *e.g.* $\mathbf{sAB}$ is the model composition that starts in state $s$, applies model $\mathbf{A}$ and then applies model $\mathbf{B}$.

### 3.1 Model Sets

Models can represent the outcomes of actions, options and policies. An option model $\mathbf{O}_o \in \mathcal{O}$ represents the outcome on termination of a corresponding option $o \in \mathcal{O}$. It combines an option transition matrix with an option reward vector, $\mathbf{O}_o = \left[\begin{array}{c|c} 1 & 0 \\ \hline R^o & P^o \end{array}\right]$. An *action model* $\mathbf{A}_a \in \mathcal{A}$ represents the outcome of a primitive action $a \in \mathcal{A}$, where $\mathbf{A}_a = \left[\begin{array}{c|c} 1 & 0 \\ \hline R^a & P^a \end{array}\right]$. Action models are option models, $\mathcal{A} \subset \mathcal{O}$, corresponding to options that terminate with probability 1. A *policy model* $\mathbf{\Pi}_\pi \in \mathbf{\Pi}$, where $\mathbf{\Pi}_\pi = \left[\begin{array}{c|c} 1 & 0 \\ \hline V^\pi & 0 \end{array}\right]$, represents the outcome of executing policy $\pi$ forever. Policy models are also option models, $\mathbf{\Pi} \subset \mathcal{O}$, corresponding to options that terminate with probability 0. Finally, we define an identity model $\mathbf{I}$ corresponding to zero reward and the identity transition matrix, $\mathbf{I} = \left[\begin{array}{c|c} 1 & 0 \\ \hline 0 & I \end{array}\right]$; this can be viewed as a *null* model without any discounting. Note that action/option/policy subscripts may be dropped when there is no ambiguity.

### 3.2 Value Models

A *value model* $\mathbf{V} \in \mathcal{V}$, where $\mathbf{V} = \left[\begin{array}{c|c} 1 & 0 \\ \hline V & 0 \end{array}\right]$, has a transition matrix of zero (*i.e.* it always exits) and a reward vector given by the components of $V(s)$ as its

reward vector (*i.e.* total reward before exiting). Policy models are value models, $\boldsymbol{\Pi} \subset \mathcal{V}$, where the reward vector contains the values $V^\pi(s)$, and the transition matrix is zero due to infinite discounting.

Value models can be used to express several familiar value functions. A state value function $V(s)$ can be represented by composition with the corresponding value model, $\mathbf{sV}$; an action value function can be represented by $\mathbf{sAV}$; and an inter-option value function (Sutton et al., 1999) can be represented by $\mathbf{sOV}$.

The *true value model* $\mathbf{G}^- = \begin{bmatrix} 1 & 0 \\ \hline V^- & 0 \end{bmatrix}$ represents the overall goal of maximising total reward. It is defined to have a value function $V^-(s)$ that is a lower bound on the value function of all policies, $V^-(s) < V^\pi(s), \forall s \in \mathcal{S}, \pi \in \Pi$. This definition ensures that a termination condition of $\beta(s) = 0$ is always optimal, and that policy models dominate over terminating option models, with respect to the true value, $\mathbf{s\Pi}_\pi \mathbf{G}^- = \mathbf{s\Pi}_\pi = \mathbf{sO}_{\langle \pi,\beta \rangle} \mathbf{\Pi}_\pi \geq \mathbf{sO}_{\langle \pi,\beta \rangle} \mathbf{G}^-, \forall s \in \mathcal{S}, \pi \in \Pi, \beta \in \mathcal{B}$.

### 3.3 Expectation Models

An *expectation model* $\mathbf{E}_\rho(\mathbf{M})$ is the expected model under some distribution $\rho(s,\cdot)$ over models. For example, an *action expectation model* $\mathbf{E}_\pi(\mathbf{A})$ averages all action models $\mathbf{A}_a \in \mathcal{A}$ according to policy $\pi(s,a)$. Specifically, each row of $\mathbf{E}_\pi(\mathbf{A})$ contains the expected rasp from state $s$ after one action has been executed by $\pi$, $E_{a \sim \pi(s,\cdot)}[\mathbf{sA}_a \mid s]$,

$$\mathbf{E}_\pi(\mathbf{A}) = \begin{bmatrix} 1 & 0 \\ \hline E_{a \sim \pi(s,\cdot)}[\mathbf{sA}_a \mid s = s_1] \\ \vdots \\ E_{a \sim \pi(s,\cdot)}[\mathbf{sA}_a \mid s = s_n] \end{bmatrix} \quad (3)$$

Composing a model with a deterministic rasp $\mathbf{s}$ picks out the row corresponding to state $s$,

$$\mathbf{sE}_\pi(\mathbf{A}) = E_{a \sim \pi(s,\cdot)}[\mathbf{sA}_a \mid s] = \sum_{a \in \mathcal{A}} \pi(s,a) \, \mathbf{sA}_a, \forall s \in \mathcal{S} \quad (4)$$

### 3.4 Maximising Models

A *max model* $\max_{\mathbf{V} \in \mathcal{W}} \mathbf{V}$ maximises over a given set of value models $\mathcal{W} \subseteq \mathcal{V}$. Each reward component is the maximum value of $\mathbf{sV}$ from state $s$.

$$\max_{\mathbf{V} \in \mathcal{W}} \mathbf{V} = \begin{bmatrix} 1 & 0 \\ \hline \max_{\mathbf{V} \in \mathcal{W}, s=s_1} \mathbf{sV} & 0 \\ \vdots \\ \max_{\mathbf{V} \in \mathcal{W}, s=s_n} \mathbf{sV} \end{bmatrix} \quad (5)$$

An *argmax model* $\mathop{\mathrm{argmax}}_{\mathbf{M} \in \mathcal{M}} \mathbf{MV}$ maximises over the models in set $\mathcal{M}$, with respect to value model $\mathbf{V}$. Each row of $\mathop{\mathrm{argmax}}_{\mathbf{M} \in \mathcal{M}} \mathbf{MV}$ is the rasp $\mathbf{sM}$ that maximises the value $\mathbf{sMV}$ from state $s$.

$$\mathop{\mathrm{argmax}}_{\mathbf{M} \in \mathcal{M}} \mathbf{MV} = \begin{bmatrix} 1 & 0 \\ \hline \mathop{\mathrm{argmax}}_{\mathbf{sM} \mid \mathbf{M} \in \mathcal{M}, s=s_1} \mathbf{sMV} \\ \vdots \\ \mathop{\mathrm{argmax}}_{\mathbf{sM} \mid \mathbf{M} \in \mathcal{M}, s=s_n} \mathbf{sMV} \end{bmatrix} \quad (6)$$

Composing an argmax model with a deterministic rasp $\mathbf{s}$ picks out the maximising row,

$$\mathbf{s} \mathop{\mathrm{argmax}}_{\mathbf{M} \in \mathcal{M}} \mathbf{MV} = \mathop{\mathrm{argmax}}_{\mathbf{sM} \mid \mathbf{M} \in \mathcal{M}} \mathbf{sMV}, \forall s \in \mathcal{S} \quad (7)$$

## 4 Model Equations

We now explore recursive relationships between compositions of models. For didactic purposes we begin with compositions of primitive actions into policy models, and develop a model equation that is analogous to the Bellman equation. We then extend this approach to compositions of option models into policy models; to compositions of action models into option models; and finally to compositions of option models into other option models. We provide proofs of unique fixed points in the supplementary material.

### 4.1 Action-Policy Model Composition

We begin by rewriting the Bellman expectation equation as a model composition,

$$\mathbf{V} = \mathbf{E}_\pi(\mathbf{A})\mathbf{V} \quad (8)$$

We call this equation the *action-policy model expectation equation*. It rewrites the Bellman expectation equation in homogeneous coordinates. This equation has fixed point $\mathbf{V} = \mathbf{\Pi}_\pi$, *i.e.* composing the action expectation model $\mathbf{E}_\pi(\mathbf{A})$ with policy model $\mathbf{\Pi}_\pi$ results in the same policy model $\mathbf{\Pi}_\pi$.

We also consider the model $\max_{\mathbf{A} \in \mathcal{A}} \mathbf{AV}$ that maximises the state-action value $\mathbf{sAV}$ from every state $s$. We can then rewrite the Bellman optimality equation in homogeneous coordinates,

$$\mathbf{V} = \max_{\mathbf{A} \in \mathcal{A}} \mathbf{AV} \quad (9)$$

We call this equation the *action-policy model optimality equation*.

The *optimal policy model* is the max model $\max_{\mathbf{\Pi} \in \Pi} \mathbf{\Pi}$ over all policy models over the set of primitive actions, $\mathbf{s} \max_{\mathbf{\Pi} \in \Pi} \mathbf{\Pi} = \max_{\mathbf{\Pi} \in \Pi} \mathbf{s}\,\mathbf{\Pi}, \forall s \in \mathcal{S}$. It is equivalent to the optimal value function. The optimal policy model $\mathbf{V} = \max_{\mathbf{\Pi} \in \Pi} \mathbf{\Pi}$ is a fixed point of the action-policy model optimality equation.

### 4.2 Option-Policy Model Composition

We now compose option models into a policy model. We assume we are given a base set of options $\Omega \subseteq \mathcal{O}$ and a corresponding set of option models $\boldsymbol{\Omega} \subseteq \boldsymbol{\mathcal{O}}$. We consider the *option expectation model* $\mathbf{E}_\pi(\mathbf{O})$ that averages the base option models $\mathbf{O}_o \in \boldsymbol{\Omega}$ according to hierarchical policy $\pi(s, o) = Pr(o \mid s)$. Similarly to Equations 3 and 4, each row of $\mathbf{E}_\pi(\mathbf{O})$ contains the expected rasp from state $s$ after one option has been executed by $\pi$,

$$\mathbf{sE}_\pi(\mathbf{O}) = E_{o \sim \pi(s, \cdot)}[\mathbf{sO}_o \mid s] = \sum_{o \in \Omega} \pi(s, o)\, \mathbf{sO}_o, \forall s \in \mathcal{S} \quad (10)$$

This gives the *option-policy model expectation equation*, with fixed point $\mathbf{V} = \boldsymbol{\Pi}_\pi$,

$$\mathbf{V} = \mathbf{E}_\pi(\mathbf{O})\mathbf{V} \quad (11)$$

Next, we consider the model $\max_{\mathbf{O} \in \boldsymbol{\Omega}} \mathbf{OV}$ that maximises the composed value $\mathbf{sOV}$ (the inter-option value function). This leads to the *option-policy model optimality equation*,

$$\mathbf{V} = \max_{\mathbf{O} \in \boldsymbol{\Omega}} \mathbf{OV} \quad (12)$$

Given only a base set of option models $\boldsymbol{\Omega}$, which does not necessarily include all primitive actions, it is not in general possible to construct *all* policy models. Instead, we consider the *hierarchical policy model set* $\{\boldsymbol{\Pi}_\pi \mid \mathrm{supp}(\pi) \subseteq \Omega\}$, which is the set of all policy models corresponding to hierarchical policies over $\Omega$. The *hierarchically optimal policy model* $\max_{\boldsymbol{\Pi}_\pi \mid \mathrm{supp}(\pi) \subseteq \Omega} \boldsymbol{\Pi}$ is the max model over this set; it is analogous to a hierarchically optimal value function (Dietterich, 2000), *i.e.* the best that can be achieved under the hierarchical constraints imposed by the choice of base options.[1] The hierarchically optimal policy model $\mathbf{V} = \max_{\boldsymbol{\Pi}_\pi \mid \mathrm{supp}(\pi) \subseteq \Omega} \boldsymbol{\Pi}$ is the unique fixed point of the option-policy model optimality equation. If the base set includes all primitive actions, $\boldsymbol{\mathcal{A}} \subseteq \boldsymbol{\Omega}$, then all policy models can be represented and the hierarchically optimal policy model is the optimal policy model, $\max_{\boldsymbol{\Pi}_\pi \mid \mathrm{supp}(\pi) \subseteq \Omega} \boldsymbol{\Pi} = \max_{\boldsymbol{\Pi} \in \boldsymbol{\Pi}} \boldsymbol{\Pi}$.

### 4.3 Action-Option Model Composition

Primitive actions can also be composed together into option models, to give intra-option model learning.

---

[1] Hierarchical optimality is a global optimality condition. In contrast, *recursive optimality* (Dietterich, 2000) is a weaker, local optimality condition that assumes all sub-options are fixed. Many hierarchical reinforcement learning algorithms achieve recursive optimality but not hierarchical optimality.

This requires a mechanism to incorporate option termination into model compositions.

We represent the termination condition $\beta(s)$ by a *termination model* $\mathbf{E}_\beta(\mathbf{I}, \mathbf{M})$. This is an expectation model over $\{\mathbf{I}, \mathbf{M}\}$ that selects each row from the identity model $\mathbf{I}$ with probability $\beta(s)$, or from model $\mathbf{M}$ with probability $1 - \beta(s)$,

$$\mathbf{sE}_\beta(\mathbf{I}, \mathbf{M}) = \mathbf{s}\left(\beta(s)\mathbf{I} + (1 - \beta(s))\mathbf{M}\right), \forall s \in S \quad (13)$$

Composing an action model $\mathbf{A}$ with termination model $\mathbf{E}_\beta(\mathbf{I}, \mathbf{M})$ selects between $\mathbf{A}$ (termination) or $\mathbf{AM}$ (continuation). In particular, we consider the composition of expectation model $\mathbf{E}_\pi(\mathbf{A})$ with termination model $\mathbf{E}_\beta(\mathbf{I}, \mathbf{M})$. This gives the *action-option model expectation equation*, with fixed point $\mathbf{M} = \mathbf{O}_{\langle \pi, \beta \rangle}$,

$$\mathbf{M} = \mathbf{E}_\pi(\mathbf{A})\mathbf{E}_\beta(\mathbf{I}, \mathbf{M}) \quad (14)$$

We now consider the optimality of option models. We define optimality with respect to a *subgoal* value model $\mathbf{G}$ that represents the value on termination of the option, *e.g.* whether a given subgoal has been achieved. An *optimal option model* $\operatorname{argmax}_{\mathbf{O} \in \boldsymbol{\mathcal{O}}} \mathbf{OG}$ is the argmax model, with respect to subgoal value model $\mathbf{G}$, over all options, *i.e.* it maximises over both policies and termination conditions. We will consider option models that maximise over policies or termination conditions in a subsequent section.

We represent optimal termination by an argmax model over $\mathbf{B} \in \{\mathbf{I}, \mathbf{M}\}$, which maximises the binary choice between termination, represented by identity model $\mathbf{I}$, and continuation, represented by model $\mathbf{M}$. For example, $\operatorname{argmax}_{\mathbf{AB} \mid \mathbf{B} \in \{\mathbf{I},\mathbf{M}\}} \mathbf{ABG}$ either selects row $s$ from action model $\mathbf{A}$ or from the composed model $\mathbf{AM}$, depending on whether $\mathbf{sAG}$ (termination) or $\mathbf{sAMG}$ (continuation) gives more reward from state $s$. We only optimise over deterministic termination conditions, because an optimal deterministic termination condition must exist (analogous to optimal policies). We can now define the *option-option model optimality equation*, for which any optimal option model $\operatorname{argmax}_{\mathbf{O} \in \boldsymbol{\mathcal{O}}} \mathbf{OG}$ is a fixed point,

$$\mathbf{M} = \operatorname*{argmax}_{\mathbf{AB} \mid \mathbf{A} \in \boldsymbol{\mathcal{A}}, \mathbf{B} \in \{\mathbf{I},\mathbf{M}\}} \mathbf{ABG} \quad (15)$$

### 4.4 Option-Option Model Composition

We now present the most general case in which option models are composed into other option models. This combines intra-option model learning with inter-option model learning, a key step towards our goal of compositional planning. As in option-policy model composition, we assume that we are given a base set $\Omega$ of options, and a corresponding set $\Omega$ of option models to compose together. As in action-option model composition, we consider termination conditions as well as policies. Combining these ideas together gives the *option-option model expectation equation*,

$$\mathbf{M} = \mathbf{E}_\pi(\mathbf{O})\mathbf{E}_\beta(\mathbf{I}, \mathbf{M}) \qquad (16)$$

with fixed point $\mathbf{M} = \mathbf{O}_{\langle \pi, \beta \rangle}$, and the *option-option model optimality equation*,

$$\mathbf{M} = \operatorname*{argmax}_{\mathbf{OB} \mid \mathbf{O} \in \Omega, \mathbf{B} \in \{\mathbf{I}, \mathbf{M}\}} \mathbf{OBG} \qquad (17)$$

| Option | Model Equation | Fixed point |
|---|---|---|
| $\langle \pi, 0 \rangle$ | $\mathbf{V} = \mathbf{E}_\pi(\mathbf{A})\mathbf{V}$ | $\mathbf{\Pi}_\pi$ |
| $\langle *, 0 \rangle$ | $\mathbf{V} = \max_{\mathbf{A} \in \mathcal{A}} \mathbf{AV}$ | $\max_{\mathbf{\Pi}_\pi \mid \pi \in \Pi} \mathbf{\Pi}$ |
| $\langle \pi, \beta \rangle$ | $\mathbf{M} = \mathbf{E}_\pi(\mathbf{A})\mathbf{E}_\beta(\mathbf{I}, \mathbf{M})$ | $\mathbf{O}_{\langle \pi, \beta \rangle}$ |
| $\langle *, \beta \rangle$ | $\mathbf{M} = \operatorname*{argmax}_{\mathbf{A}\bar{\mathbf{B}} \mid \mathbf{A} \in \mathcal{A}, \bar{\mathbf{B}} = \mathbf{E}_\beta(\mathbf{I}, \mathbf{M})} \mathbf{A}\bar{\mathbf{B}}\mathbf{G}$ | $\operatorname*{argmax}_{\mathbf{O}_{\langle \pi, \beta \rangle} \mid \pi \in \Pi, \beta = \beta} \mathbf{OG}$ |
| $\langle \pi, * \rangle$ | $\mathbf{M} = \operatorname*{argmax}_{\bar{\mathbf{A}}\mathbf{B} \mid \bar{\mathbf{A}} = \mathbf{E}_\pi(\mathbf{A}), \mathbf{B} \in \{\mathbf{I}, \mathbf{M}\}} \bar{\mathbf{A}}\mathbf{B}\mathbf{G}$ | $\operatorname*{argmax}_{\mathbf{O}_{\langle \pi, \beta \rangle} \mid \pi = \pi, \beta \in \mathcal{B}} \mathbf{OG}$ |
| $\langle *, * \rangle$ | $\mathbf{M} = \operatorname*{argmax}_{\mathbf{AB} \mid \mathbf{A} \in \mathcal{A}, \mathbf{B} \in \{\mathbf{I}, \mathbf{M}\}} \mathbf{ABG}$ | $\operatorname*{argmax}_{\mathbf{O}_{\langle \pi, \beta \rangle} \mid \pi \in \Pi, \beta \in \mathcal{B}} \mathbf{OG}$ |

Table 1. Summary of model equations and fixed points, when composing primitive action models $\mathbf{A} \in \mathcal{A}$. The *option* column indicates whether the policy $\pi$ and termination condition $\beta$ are free to be optimised ($*$), or are given.

| Opt. | Model Equation | Fixed point |
|---|---|---|
| $\langle \pi, 0 \rangle$ | $\mathbf{V} = \mathbf{E}_\pi(\mathbf{O})\mathbf{V}$ | $\mathbf{\Pi}_\pi$ |
| $\langle *, 0 \rangle$ | $\mathbf{V} = \max_{\mathbf{O} \in \Omega} \mathbf{OV}$ | $\max_{\mathbf{\Pi}_\pi \mid \operatorname{supp}(\pi) \subseteq \Omega} \mathbf{\Pi}$ |
| $\langle \pi, \beta \rangle$ | $\mathbf{M} = \mathbf{E}_\pi(\mathbf{O})\mathbf{E}_\beta(\mathbf{I}, \mathbf{M})$ | $\mathbf{O}_{\langle \pi, \beta \rangle}$ |
| $\langle *, \beta \rangle$ | $\mathbf{M} = \operatorname*{argmax}_{\mathbf{O}\bar{\mathbf{B}} \mid \mathbf{O} \in \Omega, \bar{\mathbf{B}} = \mathbf{E}_\beta(\mathbf{I}, \mathbf{M})} \mathbf{O}\bar{\mathbf{B}}\mathbf{G}$ | $\operatorname*{argmax}_{\mathbf{O}_{\langle \pi, \beta \rangle} \mid \operatorname{supp}(\pi) \subseteq \Omega, \beta = \beta} \mathbf{OG}$ |
| $\langle \pi, * \rangle$ | $\mathbf{M} = \operatorname*{argmax}_{\bar{\mathbf{O}}\mathbf{B} \mid \bar{\mathbf{O}} = \mathbf{E}_\pi(\mathbf{O}), \mathbf{B} \in \{\mathbf{I}, \mathbf{M}\}} \bar{\mathbf{O}}\mathbf{B}\mathbf{G}$ | $\operatorname*{argmax}_{\mathbf{O}_{\langle \pi, \beta \rangle} \mid \pi = \pi, \beta \in \mathcal{B}} \mathbf{OG}$ |
| $\langle *, * \rangle$ | $\mathbf{M} = \operatorname*{argmax}_{\mathbf{OB} \mid \mathbf{O} \in \Omega, \mathbf{B} \in \{\mathbf{I}, \mathbf{M}\}} \mathbf{OBG}$ | $\operatorname*{argmax}_{\mathbf{O}_{\langle \pi, \beta \rangle} \mid \operatorname{supp}(\pi) \subseteq \Omega, \beta \in \mathcal{B}} \mathbf{OG}$ |

Table 2. Summary of model equations and their fixed points, when composing option models $\mathbf{O} \in \Omega$.

It is not in general possible to construct all option models, due to limitations of the base set $\Omega$. Instead, we consider the *hierarchical option model set* $\{\mathbf{O}_{\langle \pi, \beta \rangle} \mid \operatorname{supp}(\pi) \subseteq \Omega, \beta \in \mathcal{B}\}$, which is the set of option models $\mathbf{O}_{\langle \pi, \beta \rangle}$ where $\pi$ is restricted to options in $\Omega$. The *hierarchically optimal option model*, $\operatorname*{argmax}_{\mathbf{O}_{\langle \pi, \beta \rangle} \mid \operatorname{supp}(\pi) \subseteq \Omega, \beta \in \mathcal{B}} \mathbf{OG}$, is the argmax model over this set, with respect to subgoal value model $\mathbf{G}$. A hierarchically optimal option model is a fixed point of the option-option model optimality equation.

### 4.5 Optimal $\beta$- and $\pi$-Option Models

There are in fact two dimensions of optimality for option models: optimality of the policy $\pi$ and optimality of the termination condition $\beta$. The previous sections dealt with jointly optimal option models, which maximise over both policies and termination conditions. We now consider option models that optimise just one of these two dimensions.

An *optimal $\beta$-option model* $\operatorname*{argmax}_{\mathbf{O}_{\langle \pi, \beta \rangle} \mid \pi \in \Pi, \beta = \beta} \mathbf{OG}$ is the argmax model over the set of options with termination condition $\beta$, i.e. it maximises over policies for a given termination condition $\beta$. Similarly, an *optimal $\pi$-option model* $\operatorname*{argmax}_{\mathbf{O}_{\langle \pi, \beta \rangle} \mid \pi = \pi, \beta \in \mathcal{B}} \mathbf{OG}$ is the argmax model over the set of options with policy $\pi$, i.e. it maximises over termination conditions for a given policy $\pi$.

We can now define action-option model optimality equations for optimal $\beta$-option models, where the termination condition is given; and for optimal $\pi$-option models, where the policy is given,

$$\mathbf{M} = \operatorname*{argmax}_{\mathbf{A}\bar{\mathbf{B}} \mid \mathbf{A} \in \mathcal{A}, \bar{\mathbf{B}} = \mathbf{E}_\beta(\mathbf{I}, \mathbf{M})} \mathbf{A}\bar{\mathbf{B}}\mathbf{G} \qquad (18)$$

$$\mathbf{M} = \operatorname*{argmax}_{\bar{\mathbf{A}}\mathbf{B} \mid \bar{\mathbf{A}} = \mathbf{E}_\pi(\mathbf{A}), \mathbf{B} \in \{\mathbf{I}, \mathbf{M}\}} \bar{\mathbf{A}}\mathbf{B}\mathbf{G} \qquad (19)$$

These equations have respective fixed points: optimal $\beta$-option model $\mathbf{M} = \operatorname*{argmax}_{\mathbf{O}_{\langle \pi, \beta \rangle} \mid \pi \in \Pi, \beta = \beta} \mathbf{OG}$, and optimal $\pi$-option model $\mathbf{M} = \operatorname*{argmax}_{\mathbf{O}_{\langle \pi, \beta \rangle} \mid \pi = \pi, \beta \in \mathcal{B}} \mathbf{OG}$.

For option-option model composition of $\beta$-options, we restrict option models to elements of the hierarchical option model set that also match a given termination condition $\beta$, $\{\mathbf{O}_{\langle \pi, \beta \rangle} \mid \operatorname{supp}(\pi) \subseteq \Omega, \beta = \beta\}$. The *hierarchically optimal $\beta$-option model* is the argmax model over this restricted set, $\operatorname*{argmax}_{\mathbf{O}_{\langle \pi, \beta \rangle} \mid \operatorname{supp}(\pi) \subseteq \Omega, \beta = \beta} \mathbf{OG}$.

The option-option model optimality equations for $\beta$-options and $\pi$-options respectively are,

$$\mathbf{M} = \operatorname*{argmax}_{\mathbf{O}\bar{\mathbf{B}} \mid \mathbf{O} \in \Omega, \bar{\mathbf{B}} = \mathbf{E}_\beta(\mathbf{I}, \mathbf{M})} \mathbf{O}\bar{\mathbf{B}}\mathbf{G} \qquad (20)$$

$$\mathbf{M} = \operatorname*{argmax}_{\bar{\mathbf{O}}\mathbf{B} \mid \bar{\mathbf{O}} = \mathbf{E}_\pi(\mathbf{O}), \mathbf{B} \in \{\mathbf{I}, \mathbf{M}\}} \bar{\mathbf{O}}\mathbf{B}\mathbf{G} \qquad (21)$$

The fixed points of these equations are the hierarchically optimal $\beta$-option model $\mathbf{M} = \operatorname*{argmax}_{\mathbf{O}_{\langle \pi, \beta \rangle} \mid \operatorname{supp}(\pi) \subseteq \Omega, \beta = \beta} \mathbf{OG}$; and the optimal $\pi$-option model $\mathbf{M} = \operatorname*{argmax}_{\mathbf{O}_{\langle \pi, \beta \rangle} \mid \pi = \pi, \beta \in \mathcal{B}} \mathbf{OG}$.

Table 1 and 2 summarise the various model equations and their fixed points. In the supplementary material, we prove that each fixed point satisfies the corresponding equations, and furthermore that the subgoal value of each fixed point is unique.

## 5 Option-Option Model Iteration

The Bellman optimality equation forms the basis of a wide variety of MDP planning algorithms (Sutton & Barto, 1998). Similarly, the model optimality equations can be used to derive a wide variety of MDP planning algorithms. In particular, the option-option model equations can be used to derive algorithms for compositional planning in MDPs. We focus here on a dynamic programming algorithm that uses the option-option model optimality equation (Equation 17) as an iterative update. This algorithm, which we call *option-option model iteration* (OOMI), can be viewed as a generalisation of value iteration to option models for multiple subgoals.

We assume that we are given a base set $\boldsymbol{\Omega}$ of option models, and also $m$ subgoal value models $\{\mathbf{G}_1, ..., \mathbf{G}_m\}$ for $m$ different subgoals. At each iteration $k$, the algorithm updates a set of $m$ option models $\boldsymbol{\mathcal{M}}^k = \{\mathbf{M}_1^k, ..., \mathbf{M}_m^k\}$, containing one option model for every subgoal. Each option model is initialised to the true value model, $\mathbf{M}_g^0 = \mathbf{G}^-$. At each iteration $k$, for every subgoal $j$, option model $\mathbf{M}_g^{k+1}$ is updated by the option-option model optimality equation (Equation 17). Maximisation is performed over the base set $\boldsymbol{\Omega}$ *and* the current set of option models $\boldsymbol{\mathcal{M}}^k$,

$$\mathbf{M}_g^{k+1} \leftarrow \underset{\mathbf{OB} \mid \mathbf{O} \in \boldsymbol{\Omega} \cup \boldsymbol{\mathcal{M}}^k, \mathbf{B} \in \{\mathbf{I}, \mathbf{M}_g^k\}}{\operatorname{argmax}} \mathbf{OBG}_g \qquad (22)$$

OOMI imposes no explicit hierarchy: any option model may be composed with any other option model. When updating the option model $\mathbf{M}_g$ for subgoal value model $\mathbf{G}_g$, all current option models are considered. In particular, the option model $\mathbf{M}_g$ itself is considered; this allows option models to be repeatedly squared, so that a single model may be efficiently applied as many times as required. As a result, even if OOMI is restricted to primitive actions, $\boldsymbol{\Omega} = \boldsymbol{\mathcal{A}}$, and only a single subgoal, $\mathbf{G}_1 = \mathbf{G}^-$, it may still converge in significantly fewer iterations than value iteration. We prove in the supplementary material that OOMI converges to a hierarchically optimal option model for each subgoal value model $\mathbf{G}_g$. If OOMI includes the true value model in its set of value models, $\mathbf{G}_g = \mathbf{G}^-$, then the corresponding option model $\mathbf{M}_g$ will converge to the hierarchically optimal policy model $\boldsymbol{\Pi}_{\boldsymbol{\Omega}}^*$. Option-option model iteration can similarly be extended to $\beta$-options, where the termination condition is given; or $\pi$-options, where the policy is given, by using Equations 20 and 21 respectively as iterative updates. Finally, the option-option expectation model equation (Equation 16) can be used as the basis for an iterative update, analogous to policy iteration, that interleaves option evaluation with option improvement.

## 6 Empirical Results

We illustrate our framework for compositional planning using two hierarchical MDPs: the Tower of Hanoi problem, and the *Nine Rooms* problem. The $N$-disc Tower of Hanoi problem has a discount factor is $\gamma = 1$, each action receives a reward of $-1$, and episodes terminate upon reaching the goal state ($N$ discs stacked on right peg). The level-1 Nine Rooms gridworld is a $3 \times 3$ grid. The $N$-level Nine Rooms gridworld contains a $3 \times 3$ grid of instances of level $N-1$ problems; neighbouring instances are connected by a width $3^{N-2}$ doorway; and there is a single goal state in one corner. The discount factor is $\gamma = 0.9$, rewards are 1 in the goal state, and 0 elsewhere. We also use stochastic variants in which each action causes the intended move with probability $1-p$, or with probability $p$ randomly selects another legal move (Tower of Hanoi, $p = 0.4$), or remains in the current state (Nine Rooms, $p = 0.05$). For the Tower of Hanoi, we use $m = 3N + 1$ subgoal value models. This set includes the true value model $\mathbf{G}^-$ and a subgoal value model $\mathbf{G}_{d,e} = \begin{bmatrix} 1 & 0 \\ \hline V_{d,e}^{on} & 0 \end{bmatrix}$ for placing each disc $d$ on top of each peg $e$. Each subgoal value function is defined by $V_{d,e}^{on}(s) \propto on(s, d, e)$, where the predicate $on(s, d, e)$ has a value of 1 if disc $d$ is on peg $e$ in state $s$ and 0 otherwise.[2] For the Nine Rooms, we use $12(n-1)$ subgoal value models. This set includes a subgoal value model $\mathbf{G}_{l,j}$ for each of the $j \in [1, 12]$ doorways at each level $l$ of the hierarchy. Each subgoal value function is defined by $V_{l,j}^{doorway}(s) \propto in(s, l, j)$, where the predicate $in(s, l, j)$ has a value of 1 if state $s$ is in the $j$th level-$l$ doorway, and 0 otherwise. In this problem, initiation sets were used to restrict the states considered to relevant doorways within the neighbourhood of the subgoal. All subgoal values are designed to be achieved "at any cost", by choosing a large constant of proportionality. We use the primitive actions (moving a disc in Tower of Hanoi; moving N, E, S, W in Nine Rooms) as the base set $\boldsymbol{\Omega}$.

We compare three solution methods. *Action-policy model iteration* (APMI) is a one-level planning algorithm that plans over primitive action models. It iteratively applies the action-policy model optimality equation (Equation 9), and is equivalent to value iteration. *Action-option-policy model iteration* (AOPMI) is a two-level planning algorithm, with fixed planning operators. It first performs intra-option learning, constructing option models from primitive action models by iteratively applying the action-option model optimality equation (Equation 15). It then fixes the set of option models, and performs inter-option plan-

---

[2]Results are qualitatively similar for other choices of subgoal, such as stacking or unstacking discs.

Compositional Planning Using Optimal Option Models| $N$ | APMI | | AOPMI | | OOMI | |
|---|---|---|---|---|---|---|
| | iters | backs | iters | backs | iters | backs |
| Deterministic $N$-Disc Tower of Hanoi | | | | | | |
| 1 | 2 | 2 | 2 | 4 | 2 | 6 |
| 2 | 4 | 4 | 3 | 11 | 3 | 17 |
| 3 | 8 | 8 | 4 | 23 | 4 | 29 |
| 4 | 16 | 16 | 6 | 42 | 5 | 44 |
| 5 | 32 | 32 | 12 | 76 | 6 | 62 |
| 6 | 64 | 64 | 20 | 137 | 7 | 83 |
| 7 | 128 | 128 | 40 | 253 | 8 | 107 |
| 8 | 256 | 256 | 74 | 476 | 9 | 134 |
| 9 | 512 | 512 | 148 | 914 | 10 | 164 |
| 10 | 1,024 | 1,024 | 288 | 1,779 | 11 | 197 |
| 11 | 2,048 | 2,048 | 576 | 3,499 | 12 | 233 |
| 12 | 4,096 | 4,096 | 1,142 | 6,926 | 13 | 272 |
| Stochastic $N$-Disc Tower of Hanoi | | | | | | |
| 1 | 4 | 27 | 5 | 157 | 5 | 194 |
| 2 | 12 | 162 | 13 | 626 | 8 | 291 |
| 3 | 26 | 287 | 28 | 1,285 | 14 | 780 |
| 4 | 50 | 832 | 48 | 2,640 | 22 | 1,293 |
| 5 | 98 | 1,216 | 78 | 3,972 | 30 | 2,940 |
| 6 | 193 | 2,281 | 123 | 7,193 | 38 | 4,460 |
| 7 | 383 | 6,860 | 196 | 12,752 | 46 | 5,796 |
| 8 | 763 | 10,059 | 335 | 19,200 | 54 | 6,858 |
| Deterministic level-$N$ Nine Rooms | | | | | | |
| 2 | 22 | 22 | 15 | 23 | 10 | 18 |
| 3 | 70 | 70 | 29 | 151 | 14 | 29 |
| 4 | 214 | 214 | 65 | 595 | 24 | 44 |
| Stochastic level-$N$ Nine Rooms | | | | | | |
| 2 | 24 | 24 | 41 | 67 | 22 | 52 |
| 3 | 77 | 77 | 57 | 317 | 24 | 69 |
| 4 | 239 | 239 | 95 | 980 | 33 | 90 |

Table 3. Iterations (*iters*) and backups per state (*backs*) required to solve the given problems, using action-policy model iteration (APMI), action-option-policy model iteration (AOPMI) and option-option model iteration (OOMI).

ning. Finally, it constructs a value function from option models, by iteratively applying the option-policy model optimality equation (Equation 12). *Option-option model iteration* is the compositional planning algorithm (OOMI) described in Section 5. For each algorithm we measured the total number of iterations (applications of the corresponding model equation) required; and also the mean number of backups (updates to an individual state) to each state.[3] The results are shown in Table 3. In larger instances of both problems, the compositional approach required significantly fewer iterations and backups than either flat planning (APMI) or two-level planning (AOPMI), where options are first created and then used. In the Tower of Hanoi, APMI and AOPMI required a number of iterations that grew exponentially with the number of discs, whereas OOMI required just 1 additional iteration per disc in the deterministic case, and 8 additional iterations per disc in the stochastic case. In the Nine Rooms, the total iterations for APMI and AOPMI again grows exponentially with the level, but polynomially for OOMI.

## 7 Conclusion

The Bellman optimality equation has motivated the development of a wide variety of MDP planning algorithms. We have generalised the Bellman equation in several important dimensions, enabling an analogous variety of *compositional* planning algorithms. We have illustrated one such approach, using option-option model iteration. This is the first MDP planning algorithm to dynamically create its own planning operators. These operators are composed together to give increasingly deep and purposeful jumps through state space. Like value iteration, option-option model iteration applies full-width backups over complete sweeps of the state space. In principle, the model equations could also be solved by sample backups over sample trajectories, leading to compositional algorithms for hierarchical reinforcement learning. In this paper we have focused on planning with table lookup models; however, similar to MAXQ (Dietterich, 2000), HAMs (Andre & Russell, 2002) or skills (Konidaris & Barto, 2009), substantial efficiency improvements may be generated when each option model is provided with its own state abstraction.

## References

Amarel, S. On representations of problems of reasoning about actions. *Machine Intelligence*, 3:131–171, 1968.

Andre, D. and Russell, S. State abstraction for programmable reinforcement learning agents. In *18th National Conference on Artificial Intelligence*, 2002.

Bellman, R. *Dynamic Programming*. Princeton University Press, 1957.

Dietterich, T. Hierarchical reinforcement learning with the MAXQ value function decomposition. *Journal of Artificial Intelligence Research*, 13:227–303, 2000.

Jonsson, A. The role of macros in tractable planning. *Journal of Artificial Intelligence Research*, 36:471–511, 2009.

Konidaris, G. and Barto, A. Efficient skill learning using abstraction selection. In *21st International Joint Conference on Artificial Intelligence*, 2009.

Korf, R. *Learning to solve problems by searching for macro-operators*. Pitman Publishing, 1985.

Laird, J., Rosenbloom, P., and Newell, A. Chunking in SOAR: The anatomy of a general learning mechanism. *Machine Learning*, 1(1):1146, 1986.

Parr, R. and Russell, S. Reinforcement learning with hierarchies of machines. In *Advances in Neural Information Processing Systems 10*, 1997.

Precup, D., Sutton, R., and Singh, S. Theoretical results on reinforcement learning with temporally abstract options. In *10th European Conference on Machine Learning*, 1998.

Sacerdoti, E. *A structure for plans and behavior*. PhD thesis, Stanford University, 1975.

Sutton, R. TD models: Modeling the world at a mixture of time scales. In *12th International Conference on Machine Learning*, pp. 531–539, 1995.

Sutton, R. and Barto, A. *Reinforcement Learning: an Introduction*. MIT Press, 1998.

Sutton, R., Precup, D., and Singh, S. Between MDPs and semi-MDPs: A framework for temporal abstraction in reinforcement learning. *Artificial Intelligence*, 112(1-2):181–211, 1999.---

[3] Note, however, that each backup has a larger cost with model iteration, since a complete row must be updated.